\def\year{2020}\relax
\documentclass[letterpaper]{article} % DO NOT CHANGE THIS
\usepackage{aaai20}  % DO NOT CHANGE THIS
\usepackage{times}  % DO NOT CHANGE THIS
\usepackage{helvet} % DO NOT CHANGE THIS
\usepackage{courier}  % DO NOT CHANGE THIS
\usepackage[hyphens]{url}  % DO NOT CHANGE THIS
\usepackage{graphicx} % DO NOT CHANGE THIS
\usepackage{graphicx}  % DO NOT CHANGE THIS
\graphicspath{ {} }
\usepackage{amsmath,amssymb}
\title{FLNet: Landmark Driven Fetching and Learning Network for Faithful Talking Facial Animation Synthesis}
\DeclareMathOperator{\E}{\mathbb{E}}
\DeclareMathOperator{\R}{\mathbb{R}}
\author{Kuangxiao Gu, Yuqian Zhou\thanks{Corresponding author}, Thomas Huang\\ % All authors must be in the same font size and format. Use \Large and \textbf to achieve this result when breaking a line
Image Formation and Processing Group,\\University of Illinois at Urbana-Champaign, Urbana, IL, USA, 61801\\
\{kgu3, yuqian2\}@illinois.edu%If you have multiple authors and multiple affiliations
}

\begin{document}

\maketitle

\begin{abstract}
Talking face synthesis has been widely studied in either appearance-based or warping-based methods. Previous works mostly utilize single face image as a source, and generate novel facial animations by merging other person's facial features. However, some facial regions like eyes or teeth, which may be hidden in the source image, can not be synthesized faithfully and stably. In this paper, We present a landmark driven two-stream network to generate faithful talking facial animation, in which more facial details are created, preserved and transferred from multiple source images instead of a single one. Specifically, we propose a network consisting of a learning and fetching stream. The fetching sub-net directly learns to attentively warp and merge facial regions from five source images of distinctive landmarks, while the learning pipeline renders facial organs from the training face space to compensate. Compared to baseline algorithms, extensive experiments demonstrate that the proposed method achieves a higher performance both quantitatively and qualitatively. Codes are at \url{https://github.com/kgu3/FLNet_AAAI2020}.
\end{abstract}

\section{Introduction}
Facial animation generation has been an active research for decades. Being able to generate realistic talking facial animation is a crucial step in human computer interaction. Studies \cite{Azevedo2018UsingAdults.} show that for healthcare services, a face-to-face chat setting actually helps promote the effectiveness of communication and leads to better patient outcomes. A robust talking face synthetic system can also benefit video compression and transmission for teleconference systems or electronic healthcare systems. It also has wide applications in entertainment and animation industry. 

Traditional methods \cite{Xie2007RealisticModelling,Blanz2003ReanimatingVideo,Yu2012Perception-drivenSynthesis,Blanz1999AFaces} mainly focus on either building a statistical face representation using techniques like principal component analysis (PCA), or constructing a 3D face model with textures from an example image. Due to limited representation and generalization ability, few of them can achieve robust and satisfactory performance. 

The recent advancements in generative adversarial networks (GANs) \cite{NIPS2014_5423} enable researchers to generate much more realistic facial animations \cite{Pham2018GenerativeNetwork,Pumarola2018GANimation:Image,Geng2018Warp-GuidedAnimation,Song2017GeometrySynthesis,zhou2017photorealistic}. Such methods utilize large-scale training data and high capacity neural networks to directly generate facial images from different input modalities, such as audio, video, facial landmark or semantic labels like Facial Action Units (AUs). However, the generated result often suffers from artifacts such as blurriness, missing or mismatching of facial details. Some facial characteristics of the subjects like tooth, eyeballs, lip shape, or wrinkles are synthesized inaccurately with pure appearance-based or geometric-based methods.  

To render more faithful facial details which are consistent with the source subject, one intuitive idea is to utilize more face images of the person. It is feasible because of two aspects. First, obtaining or collecting multiple instances of a subject is not a hard task in practical applications. Second, for the task of synthesizing talking faces, no extreme head motions or expression changes are required to model or present, while great efforts should be focused on generating faithful facial details like eyes and mouth where humans are especially sensitive with. Therefore, in this work, we propose to combine the warping-based and appearance-based methods and infer through multiple source images of the subject.

Specifically, we decouple the synthesis network into two streams: warping-based fetching and appearance-based learning. In the fetching stream, we select specific facial regions from the source image bank, and warp then merge them to form the target face. In the learning stream, a pure appearance-based method is applied to learn the facial appearance feature space from the training dataset. Finally, we merge the two streams' outputs to obtain the final result. These two streams will complement each other. Fetching stream extracts useful facial characteristics from the input source images to warp, and learning stream compensates to generate other unseen appearance features which does not exist on the source images. By combining the advantages of the appearance-based learning and warping-based fetching streams, our method generates a smooth animation with highly preserved speaker identity and facial details.

In summary, the main contributions of the paper follow,
\begin{itemize}

\item We propose an end-to-end system combining appearance-based and warping-based methods for talking facial animation generation. The final result does not require any post-processing.

\item Instead of using single image as the source, we utilize an image bank with five source images of distinctive facial landmarks from the same subject. We claim that, due to the easy accessibility of the data, the method is efficient to synthesize faithful and realistic facial details which are consistent with the target subject.
\end{itemize}

\section{Related Works}

\paragraph{Facial Action Units (AUs) Based Methods}
One approach is to synthesize facial animation based on action unit (AU). Pham et al. \cite{Pham2018GenerativeNetwork} used an encoder to extract identity information, which is concatenated with AU and fed into decoder to generate target face image. A discriminator along with a identity classifier are used to get better image quality while at the same time preserve face identity. In addition, an AU estimator is used as additional loss to make sure the decoder generates face image with desired expressions. 

Similarly, \cite{Pumarola2018GANimation:Image} used AU based approach. Instead of generating the face image, they generate the color change between input image and target image as well as a merging mask. The color change and input image are weighted by the merging mask and merged together to get the final output. Compared to \cite{Pham2018GenerativeNetwork}, this approach utilizes the vast amount of information already available in the input image, such that the generator can focus on the difference between input and output, which is similar to residual learning applied on a whole network level. Since the generator does not need to learn to generate the whole face, it can focus on the important part of face where most changes happen. As a result, the generated image is both sharp and realistic. Despite the fact that AU is intuitive to use, it is a high-level face descriptor and cannot fully model the fine detailed changes in facial geometry. In addition, their method did not use any warping when generating the target face image, which limits their generation capability. In the case where the target and source faces are not aligned, their method failed to generate sharp image. 

\paragraph{Landmark Driven Methods}
Landmark guided approach has been studied by various researchers. \cite{Song2017GeometrySynthesis} used landmark heatmap as input to replace AU, which can better guide the facial structure generation process with more detail. Their network takes a U-Net structure, with heatmap and identity face image as input and target image as output. Still, their approach did not use any warping method and did not address the hidden region problem. 

On the other hand, \cite{Geng2018Warp-GuidedAnimation} used warping-based method, which transfers the expression from a source face image to the target identity image by a sequential process, involving 3D face mesh estimation, landmark re-targeting, coarse warping, GAN refinement and tooth region hallucination. Their approach was able to generate highly realistic target image, but the generation process is lengthy and requires many steps. In addition, the tooth region is hallucinated, meaning that there is no guarantee the generated tooth region will match the actual tooth of the person, which will lead to lower authenticity. X2Face \cite{WilesX2FaceCodes} also used a warping based approach, which starts by mapping multiple input face images to an embedded face image (essentially a frontalized face image), then use a driving image to generate a dense pixel-wise warp field, which will be used to warp the embedded face to get the target face. Their approach was able to take advantage of the information in the source image, but since the embedded face image is only just one image, it lacks variation in lip shape and tooth region, which will have a negative impact on generating unseen tooth region. Similar to landmark heatmap, facial edge has also been used to guide the generation process. Specifically, \cite{ZakharovFew-ShotModels} used face edge map and adaptive instance normalization(AdaIN)\cite{Huang2017ArbitraryNormalization} during the generation process, where the parameters of AdaIN was regressed from face image embedding. \cite{WangVideo-to-Video2018} used landmark connected lines to generate high resolution talking face videos in a sequential frame-by-frame approach. Additionally, works on facial video retargeting \cite{bansal2018recycle} and facial appearance transfer \cite{zhao2019look} have been studied and remarkable results have been achieved.

Our work combines the advantages of appearance-based and warping-based methods. However simply stitching those two together would not solve the problem of hidden regions like tooth or eyelid. In order to get a high-quality output image, the strength of each method needs to be combined with a mechanism where more visual information can be extracted and coherently combined. We achieved this by using automatically learned attention masks to extract facial features from different source images such that the network can generate novel output images with highly preserved details.

\section{Method}
\subsection{Network Structure}
\begin{figure*}[ht]
    \centering
    \includegraphics[width=\textwidth]{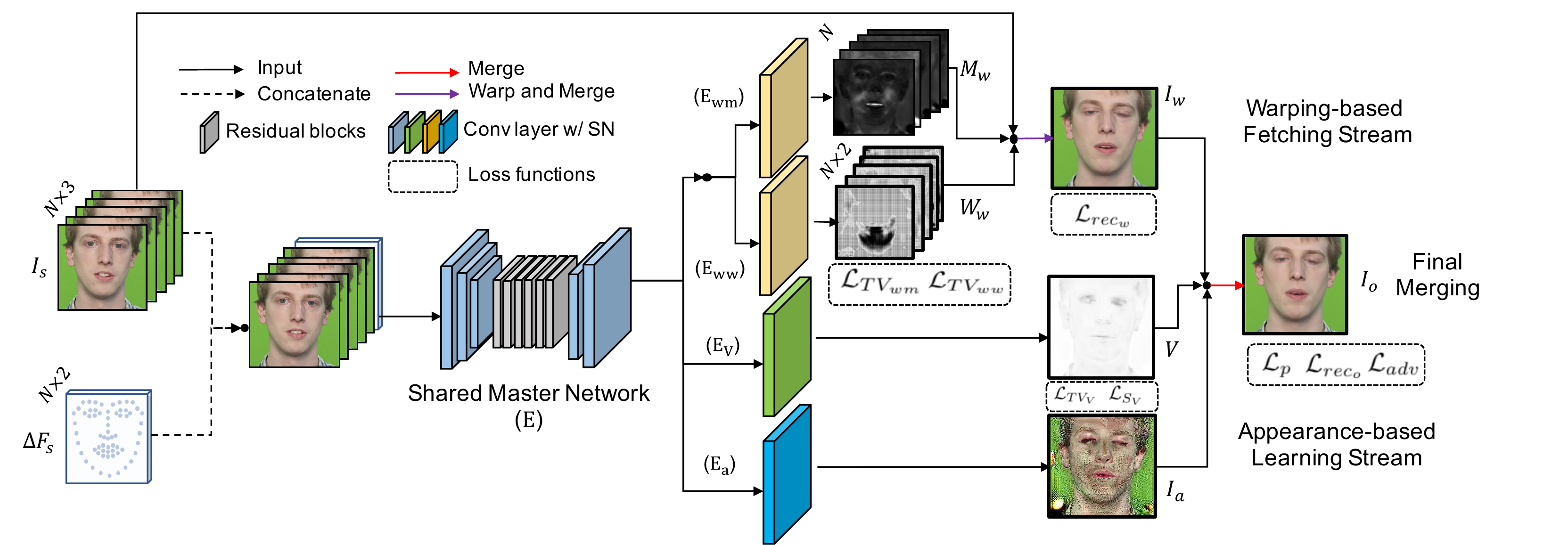}
    \caption{Network structure. The model mainly consists of a warping-based fetching stream and another appearance-based learning stream. Then the outputs of the two streams, $I_w$ and $I_a$, will be merged together by a learned selection mask $V$. The inputs of the network will first go through a shared master network $E$ following the structure of GANimation \cite{Pumarola2018GANimation:Image}, and then split into different branches $E_a$, $E_{wm}$, $E_{ww}$ and $E_V$ which are all single-layer 2D Convolution.  }
    \label{fig:net_struct}
\end{figure*}
The network structure is illustrated in Figure \ref{fig:net_struct}. The model mainly consists of a warping-based fetching stream and another appearance-based learning stream. Then the outputs of the two streams will be merged together by a learned selection mask. The inputs of the network will first go through a shared master network $E$ following the structure of GANimation \cite{Pumarola2018GANimation:Image}, and then split into different branches $E_a$, $E_{wm}$, $E_{ww}$ and $E_V$ which are all single-layer 2D Convolution to be introduced later.  

\subsubsection{Network Inputs}
The inputs of the network consist of a source image bank $I_s$, and the facial landmark difference $\Delta F_s$ between target and source faces. 

For the source image bank $I_s$, we form it by automatically selecting $N$ face images based on the mouse openness. The images are selected from the videos or frames which are different from the ones we use to extract the target landmark sequence. We sort all the frames in a sequence such that the first image has closed mouth and the last one has opened mouth. Then we uniformly select $N$ frames in order from them. To measure the mouse openness, we use the distance between outer middle lip landmarks (index 52 \& 58 for Dlib \cite{King2009Dlib-ml:Toolkit} landmark staring from 1) as criteria for selection. 
We then crop the face region and place it in the center of the image.

Different from common approaches, we did not perform heavy alignment on the face image for two reasons. First, the aligned face images often show zoom-in, zoom-out effect and instability outside the face region, which would make the warping step later much more difficult. Second, we would like the network to learn the spontaneous head movement during talking, so that the network will be able to generate such movement, which makes the generated animation much more vivid and realistic.

Given $N$ input images in the source bank $I_{s} = \{x_i\}$ where $x_i \in \R^{W \times H \times 3}$, $i=1,...,N$ and $W, H$ is the width and height of the image, we extract the corresponding 68 facial landmarks for each image $x_i$ as $S_i = \{s_j\}$ where $s_j \in \R^{2}$ and $j=1,...,68$. Given the target facial landmark $t_j \in \R^2$, $j=1,...,68$, we compute the target-source difference coordinate sets as $\Delta S_i = \{t_j-s_j\}$ for each $x_i$. Then we represent the facial landmark difference $\Delta F_s = \{ f_i \}$ where $f_i \in R^{W \times H \times 2}$ as spatial field maps by filling the positions indicated by $S_i$ with values in $\Delta S_i$ and elsewhere zero. Then we concatenate the tensor $I_s$ of size $(W, H, N \times 3)$ with $\Delta F_s$ of dimension $(W, H, N \times 2)$ to form the input tensor of dimension $(W, H, N \times 3 + N \times 2)$. The master encoder-decoder network $E$ will generate the shared feature $I_f$ by,
\begin{equation}
    I_f = E(I_s, \Delta F_s; \theta),
\end{equation}
where $\theta$ is the network parameters.
Compared with other landmark encoding methods like spatially expanding the landmark locations and channel-wisely concatenating with input images, our encoding method yields less blur generated image and a smaller input tensor. Compared with heatmap encoding \cite{Song2017GeometrySynthesis,Dong2018Soft-GatedSynthesis,SiarohinDeformableGeneration,Fragkiadaki2015RecurrentDynamics,Jackson2017LargeRegression}, our proposed method requires less computation.

\subsubsection{Warping-based Fetching Stream}
The warping-based fetching stream is deployed to fully reuse the appearance features in the source image bank $I_s$ from the same subject by warping and merging them. The stream outputs the selection mask set $M_w = \{ m_i \}$ where $m_i \in \R^{W \times H \times 1}$, and the warping field set $W_w = \{ w_i \}$ where $w_i \in \R^{W \times H \times 2}$ and $i=1,...,N$. The dimension number 2 stands for horizontal and vertical distances of the warping field. Here we represent $M_w$ and $W_w$ in tensors of size $(W, H, N)$ and $(W, H, N \times 2)$, then
\begin{equation}
    M_w =  softmax(E_{wm}(I_f;\theta_{wm}))
\end{equation}
\begin{equation}
    W_w =  E_{ww}(I_f;\theta_{ww})
\end{equation}

For the selection mask, $m_i$ takes values in range $[0,1]$ after softmax so $\sum_{i=1}^N{m_i}=J_{W, H}$ where $J$ is the all-ones matrix. The intensity of $m_i$ will be higher if the landmarks of $x_i$ better match the target landmarks. For the warping field $w_i$, instead of directly generating the warping field in terms of pixel unit, we generate the warping field in range $[-1,1]$ to specify the limits of warping distance by $[-M, M]$ either horizontally or vertically. $M$ is a pre-defined constant warping margin set as 40. This allows for flexible control during the generation process, where we can change $M$ value to control the strength of warping and the amount of motion in the generated images.

Then the warped output $I_w$ will be,
\begin{equation}
    I_{w}^c = \sum_{i=1}^N m_i \odot \phi(x_i^c,w_i), 
\end{equation}
where $\phi(\cdot)$ is the bi-linear warping function, and $\odot$ is the element-wise product and c is the RGB channel index.
\subsubsection{Appearance-based Learning Stream}
The appearance-based learning stream is trained to model the facial feature space, and complement the unseen facial textures not existing in the source image bank $I_s$. Although multiple images are fed as input, some features and pixels may be still missing for warping, and requires additional synthesis from external resources. For example, when blinking, the upper eyelid region, which is very unlikely to be captured in the input image, will be synthesized in the learning stream output. There are some other cases like the wrinkle around mouth region, the color change between teeth (depending on mouth openness) as well as the shadow under jaw when the mouth is opened. Similar to residual learning, the appearance-based branch forces the network to learn the difference between input and output image, allowing it to focus on the important task of generating unseen pixels. The appearance-based branch takes the feature $I_f$ from the master network $E$ and pass them through a single layer 2D Convolution layer $E_a$ to generate an RGB image,
\begin{equation}
    I_a = E_a(I_f;\theta_a).
\end{equation}

\subsubsection{Final Result Merging}
To merge $I_a$ and $I_w$ generated by the above two streams, the network will first learn another single-channel merging mask $V \in \R^{W \times H \times 1}$, 
\begin{equation}
    V = \sigma(E_{V}(I_f;\theta_{v})).
\end{equation}
Here $\sigma$ is the sigmoid operation. Each pixel in $V$ ranges $[0,1]$, where 0 means the pixel in the final output image $I_o$ is selected from $I_{a}$, and 1 means the pixel in the final output image is selected from the warped image $I_{w}$.

The final output image $I_o \in \R^{W \times H \times 3}$ is then simply generated by,
\begin{equation}
    I_{o}^c = (J_{W, H}-V) \odot I_{a}^c + V \odot I_{w}^c.
\end{equation}

\subsection{Objective Functions}
Following \cite{Pumarola2018GANimation:Image}, we impose a sparsity loss $\mathcal{L}_{S_{V}}$ to the merging mask $V$ to encourage the network to take some information from the appearance branch.
\begin{equation}
    \mathcal{L}_{S_{V}} = \left\lVert V \right\rVert_1
\end{equation}
We do not apply the same loss to $M_w$ because in an extreme case, if the input image is perfectly matched to the target landmark, then the network can simply choose that input image as a trivial solution, in which case the mask can be all ones and not sparse. 

We further introduce the total variation loss to $V$, $M_w$ and $W_w$ to enhance the smoothness of the learned maps. Specifically,
\begin{equation}
    \mathcal{L}_{TV_V} = \left\lVert \bigtriangledown_h V \right\rVert_2^2 + \left\lVert \bigtriangledown_v V \right\rVert_2^2
\end{equation}
\begin{equation}
    \mathcal{L}_{TV_{wm}} = \left\lVert \bigtriangledown_h M_w \right\rVert_2^2 + \left\lVert \bigtriangledown_v M_w \right\rVert_2^2
\end{equation}
\begin{equation}
    \mathcal{L}_{TV_{ww}} = \left\lVert \bigtriangledown_h W_w \right\rVert_2^2 + \left\lVert \bigtriangledown_v W_w \right\rVert_2^2
\end{equation}

To achieve a better reconstruction performance, we introduce a weighted L1 loss to $I_w$ and $I_o$. We impose L1 loss on $I_w$ in order to mitigate the effect of errors from warped result on other parts of the network by direct supervision. Without L1 loss on $I_w$, the warped result was blur and the network tended to rectify that through the appearance branch. The overall result is then over smoothed. The weight matrix $K$ is basically a heatmap where regions around landmarks have higher weights while regions far from landmark have lower weights. $K=\gamma ^{B}$ where $B \in \R^{W \times H}$ is the matrix recording the distance between each pixel and the nearest landmark of the target face $I_t$. $\gamma$ is empirically set to 0.95. Finally we lower bound the values in heatmap to 0.3 so that the network can take into account the backgrounds as well.
%the nearest landmark
\begin{equation}
    \mathcal{L}_{rec_w} = \sum_{c=1}^3\left\lVert K \odot (I_w^c - I_t^c) \right\rVert_1,
\end{equation}
\begin{equation}
    \mathcal{L}_{rec_o} = \sum_{c=1}^3\left\lVert K \odot (I_o^c - I_t^c) \right\rVert_1.
\end{equation}
The total reconstruction loss will be,
\begin{equation}
    \mathcal{L}_{rec} = \mathcal{L}_{rec_w}+\mathcal{L}_{rec_o}.
\end{equation}

We also use perceptual loss \cite{Johnson2016PerceptualSuper-resolution} on $I_o$ to preserve fine details. 
\begin{equation}
    \mathcal{L}_{p} = \left\lVert VGG(I_o) - VGG(I_t)\right\rVert_2^2,
\end{equation}
where $VGG$ is the VGG16 network which is same as \cite{Johnson2016PerceptualSuper-resolution}. We take relu1\_2, relu2\_2, relu3\_3 and relu4\_3 layers to extract high-level features.

We also apply adversarial loss as in \cite{Pumarola2018GANimation:Image} and PatchGAN \cite{IsolaImage-to-ImageNetworks} to improve the quality of the final output image. We use hinged version of Spectral Normalization(SN) \cite{TakeruMiyatoToshikiKataokaMasanoriKoyama2018SpectralNetworks} which was shown to improve the generation quality. In addition, we follow the approach in \cite{Shrivastava2017LearningTraining} to update the discriminator $D$, following the same structure as \cite{Pumarola2018GANimation:Image} except for the SN layer, using a history of generated fake images, which improved the training stability significantly. The adversarial loss is defined as,
\begin{align}
    \begin{split}
        \mathcal{L}_{adv} &= \E_{x\sim P_{real}}[\max{(0,1-D(x))}] \\
                &+ \E_{(I_s,\Delta F_s)\sim P_{in}}[\max{(0,1+D(G(I_s, \Delta F_s; \theta_{all})))}]
    \end{split}
\end{align}
where $P_{real}$ is the ground truth data distribution and $P_{in}$ represents the data distribution of the input space. $G$ is the overall generator structure and $\theta_{all}$ is the combined trainable parameters.

Finally, the full loss is a weighted sum of each individual loss:
\begin{equation}
\begin{split}
    \mathcal{L} &=\lambda_S \mathcal{L_{S_V}} \\
         &+\lambda_{TV} (\mathcal{L}_{TV_V} + \mathcal{L}_{TV_{ww}}+ 0.1\times \mathcal{L}_{TV_{wm}} ) \\
         &+\lambda_{rec} \mathcal{L}_{rec} + \lambda_{p}\mathcal{L}_{p} + \lambda_{adv} \mathcal{L}_{adv}
\end{split}
\end{equation}

\section{Experiment}
\subsection{Dataset}
We use TCD-TIMIT \cite{Gillen2014TCD-TIMITRecognition} and FaceForensics \cite{Rossler2018FaceForensics:Faces} datasets.
\paragraph{TCD-TIMIT.} TCD-TIMIT contains 62 speakers with totally 6913 videos. We choose this dataset because all the videos are in high resolution that capture the fine details of human faces including the mouth region and tooth. It enables us to better evaluate the details and identity preservation capability. We use the suggested train/test split provided by the dataset, where training and testing speakers are disjoint. 
\paragraph{FaceForensics.} The FaceForensics contains 1004 facial region videos acquired from YouTube. Although this dataset has lower resolution on face region compared to TCD-TIMIT, it has much larger color, illumination, expression and background diversity. We only use the real videos in this dataset in our experiment. The train/test was split randomly into 75\% and 25\% without overlapping.\\

To prepare the training and testing data, we first detect landmarks in each frame using Dlib \cite{King2009Dlib-ml:Toolkit}. For TCD-TIMIT, we crop the face region and resize to $224\times224$ with nose tip placed at the center. No further alignment was performed. For FaceForensics, the nose center landmark was less stable than TCD-TIMIT due to its lower resolution. As a result, we center the face images using the two outer eye corners, which is more robust and the cropped face region is more stable.

During training, for TCD-TIMIT dataset we randomly select 16 frames from 4 videos as target images. The $I_{s}$ was randomly selected from other videos of the same person different from the ones used to get target images. For FaceForensics dataset, $I_{s}$ was selected from the same video since each identity in this dataset has only one video, but we make sure that the target images are different from source images $I_{s}$.

During testing, we extract the $I_{s}$ and target landmarks $t$ using the same protocol as the training phase. Then we generate the output face images frame by frame. Finally, we simply accumulate the generated image sequence to obtain the video. No any form of post processing was performed.

\subsection{Implementation}
We implemented the system using PyTorch. Our shared master network $E$ follows the structure in \cite{Pumarola2018GANimation:Image}. It consists of three convolutional layers which reduce the input spatial dimension by a factor of 4 while increasing the number of channels to 256. Six residual blocks are used to further process the features on the bottleneck. Then two convolution layers bring the feature back to original spatial resolution while reducing channel from 256 to 64. After that, the network branches into 4 sub-networks, where each sub network contains only one layer of convolution layer before generating the corresponding outputs. We use instance normalization in the generator and spectral normalization (SN) \cite{TakeruMiyatoToshikiKataokaMasanoriKoyama2018SpectralNetworks} after each convolution layer. The activation function is ReLU. The whole network is fully convolutional, so it can handle different input sizes.

The discriminator has four convolutional layers which reduce the input spatial dimension by a factor of $2^4$. We used spectral normalization after each convolution layer. The output of the last convolution layer is used as patchGAN output. No instance or batch normalization is used in discriminator.

For every 5 iterations of discriminator update, we train once the generator. We used Adam with $(\beta_1,\beta_2)=(0.5,0.999)$ and learning rate $1\mathrm{e}{-4}$ for both the generator and the discriminator. The coefficients for the total loss are $(\lambda_{S_V},\lambda_{TV},\lambda_{rec},\lambda_{p},\lambda_{adv} )=(1.0,1\mathrm{e}{-5},250,1.0,1.0)$, which are selected empirically.

\subsection{Baseline Methods}
We choose two baseline algorithms to evaluate and compare our approach: an appearance-based GANimation~\cite{Pumarola2018GANimation:Image} and a warping-based X2Face~\cite{WilesX2FaceCodes}.
\subsubsection{GANimation.} The original GANimation used action units (AU) as conditional input for facial expression editing, which was not able to model fine changes around the mouth region, or head rotations. Landmarks contain much richer information of facial actions than AUs. As a result, we slightly modify it by replacing AU label with target landmarks as the conditional input, while everything else are kept unchanged. It would make a fair comparison between modified GANimation and our method. We used the code provided by the author and trained it on TCD-TIMIT dataset and FaceForensics separately. During testing, we extracted the target landmarks from the ground truth driving images along with an identity image and fed into GANimation to get the synthesized image.
\subsubsection{X2Face.} For fair comparison, We used the provided code and trained it on the same two datasets. During testing, since X2Face can use multiple identities to get the embedded face image, we use 5 images containing different mouth shapes as in our method. This ensures a fair comparison by supplying both methods the same amount of information as input. 

We follow the original training protocols in each models to obtain the best performance of the two models.

\subsection{Qualitative evaluation}
\begin{figure*}[ht]
    \centering
    \includegraphics[width=\textwidth]{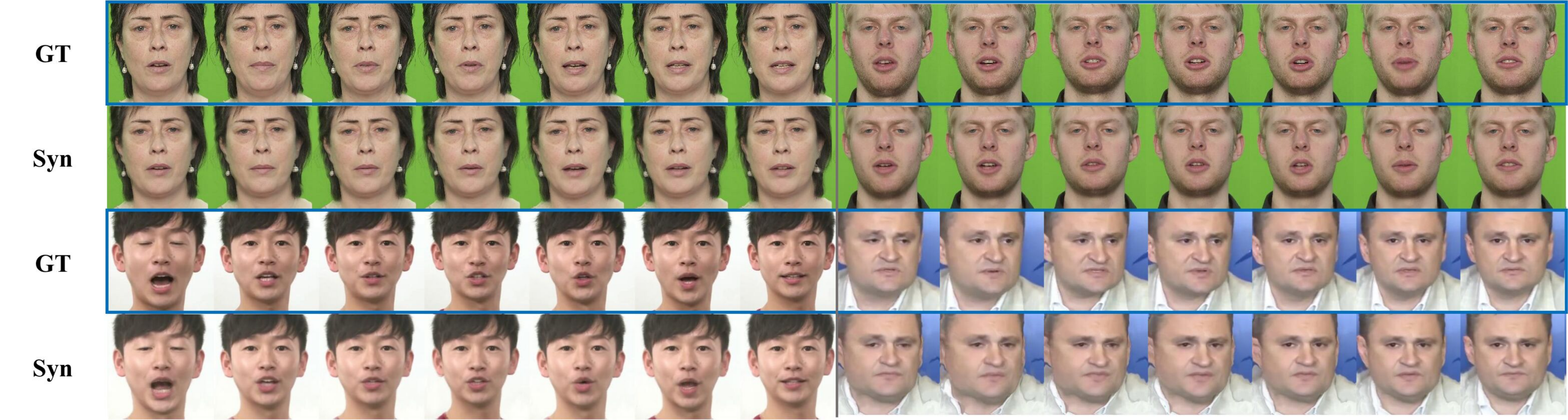}
    \caption{Synthesis result on both datasets. Upper row: ground truth. Bottom row: synthesized result.}
    \label{fig:output}
\end{figure*}

Figure \ref{fig:output} demonstrates our generated images, compared with the ground truth. Note that our method is able to preserve fine details that closely match the ground truth person identity, especially the mouth region and tooth shape. In addition, the temporal consistency is well addressed without any post processing involved.

\begin{figure*}[ht]
    \centering
    \includegraphics[width=\textwidth]{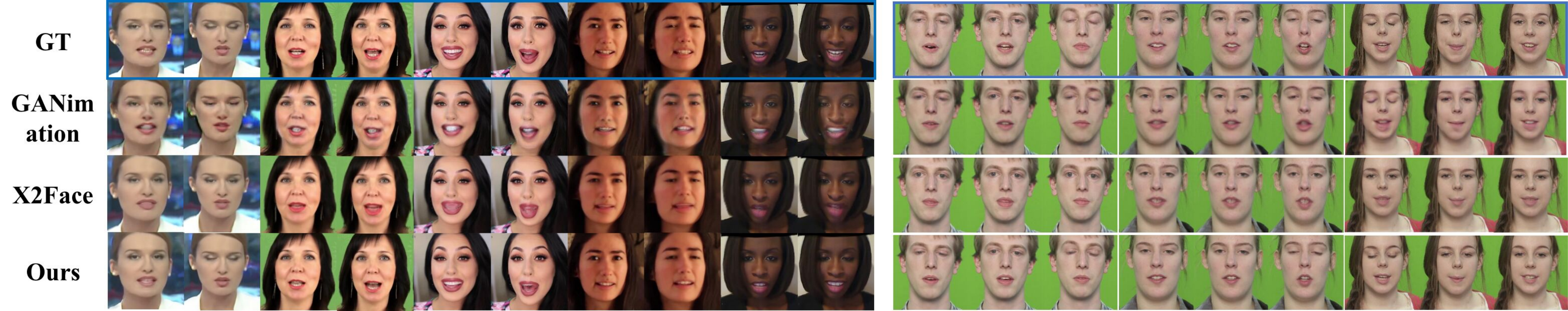}
    \caption{Comparisons with GANimation and X2Face on FaceForensics(left) and TCD-TIMIT(right). Note that X2Face was not able to generate closed eye image. GANimation was able to generate hidden region like eyelid and tooth from its color change branch, but struggled at generating regions with very fine details like lips and tooth. Our method combines the advantages of the two models by preserving fine details of the face and compensating more details not existing in the source bank images.}
    \label{fig:compare}
\end{figure*}

Figure \ref{fig:compare} compares our method with GANimation and X2Face. GANimation is not able to authentically, faithfully, and consistently reconstruct the mouth region, especially for opened mouth where tooth becomes visible. That's because only one image was used as identity input to present the similar visual cues. It yields blur results by sampling from other training samples. Compared to GANimation, X2Face was able to use multiple images to construct the embedded face image, which contains more information than the single image input method that GANimation used. However, the embedded face is still just one image, which can only provide limited facial detail and variation. Note that the X2Face result better preserved mouth region and facial detail than GANimation while still suffers from blurring effect. It also fails when some other cues like closing eyes do not exist in the source image bank. In comparison, our method is able to generate face images visually much better with fine details that are closely matched to the ground truth image. That is because we combine the advantages of the warping-based and appearance-based methods, by both fully using the facial features in the source banks, and compensating other unseen features using adversarial appearance-based learning.

\subsection{Quantitative evaluation}
We evaluated our model quantitatively using L1 loss and FID score \cite{Heusel2017GANsEquilibrium}. The L1 loss measures the absolute pixel value error, which is good at quantifying the absolute distance between reconstructed face image and the ground truth. FID score, on the other hand, measures the perceived error and quality of generated image, compared to ground truth image, which better mimics human perception.

The quantitative result is shown in Table \ref{tab:compare}. As we can see that our proposed method achieved a better result in both FID score and L1 loss. Our method was able to get the best score due to the combined advantage of combined appearance and warping streams.
\begin{table}[t]\setlength{\tabcolsep}{7pt}
\centering
\begin{tabular}{l|c|c|c|c}
\hline
           & \multicolumn{2}{c|}{TIMIT} & \multicolumn{2}{c}{FaceForensics} \\ \hline
           & L1          & FID         & L1              & FID             \\ \hline
GANimation\shortcite{Pumarola2018GANimation:Image} & 10.86       & 59.65       & 16.19           & 47.99           \\ \hline
X2Face\shortcite{WilesX2FaceCodes}     & 8.31        & 30.50       & 11.05           & 23.98           \\ \hline
Ours       &\textbf{7.99} &\textbf{17.07} &\textbf{10.20} &\textbf{20.62}  \\ \hline        
\end{tabular}

\caption{Quantitative comparison. Both L1 and FID score are lower the better. Our method achieved the best performance in both FID score and L1 loss on both datasets.}
\label{tab:compare}
\end{table}

\subsection{Ablation study}
While it has been shown in GANimation that the attention mask and color change improved the generation quality, the effectiveness and necessity of warping mechanism for high fidelity facial image generation remains unanswered. To address this question, we preformed ablation study by removing the warping or appearance stream respectively. In addition, we study the impact of the variation in the source images bank $I_s$ on the generation quality. Specifically, the first 1, 3 or all 5 images are put in the source image bank $I_s$ to evaluate the synthesis performance. 

\begin{figure}[ht]
    \centering
    \includegraphics[width=1\columnwidth]{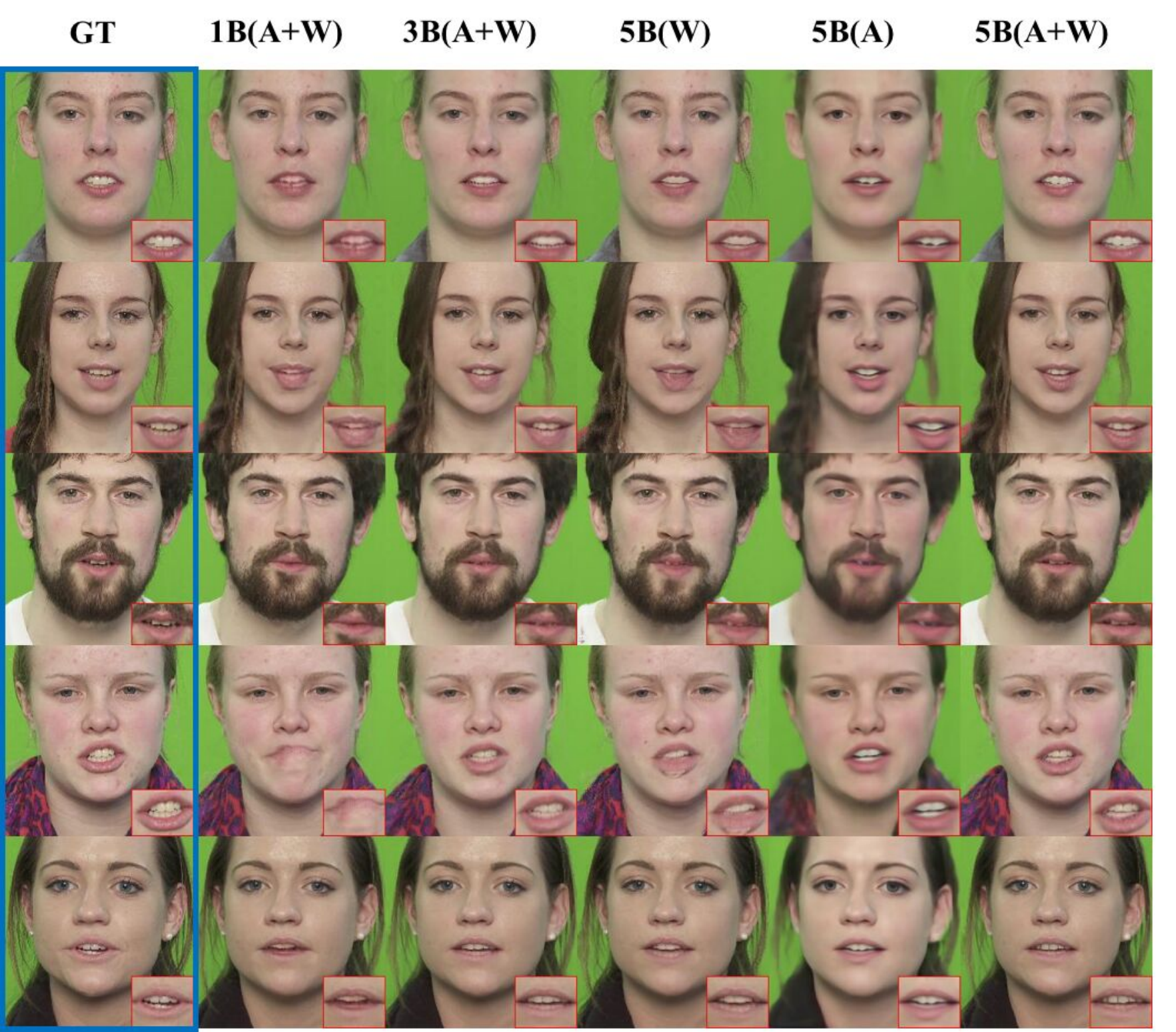}
    \caption{Ablation study. Single image as input source cannot faithfully generate the details. The three images version yields better result. The appearance-based stream only network generates plausible results, but the details are blurred out. The warping-based stream only variant suffers from the same hidden region problem. The full model is able to generate detailed facial image that closely matches the ground truth. GT: ground truth image. 1B:one image as source, 3B:three images as source, 5B:all five images as source, A:appearance-based stream only, W:warping-based stream only, A+W:(full model). The red square in each image shows the zoomed mouth region.}
    \label{fig:ablation}
\end{figure}

As we can see in Figure \ref{fig:ablation}, single image as input source cannot faithfully generate the details on the face especially for the hidden region. The same problem is observed on the other two comparison methods as well (see Figure \ref{fig:compare}). The three images version yields better result, but still lacks detail. The appearance-based stream only network generates plausible results, but the details like hair, tooth and wrinkles are blurred out. The warping-based stream only variation suffers from the same hidden region problem. The full model is able to generate detailed facial image that closely matches the ground truth.

\subsection{Visualizations of Selection Masks}
\begin{figure}[ht]
\centering
\includegraphics[width=\columnwidth]{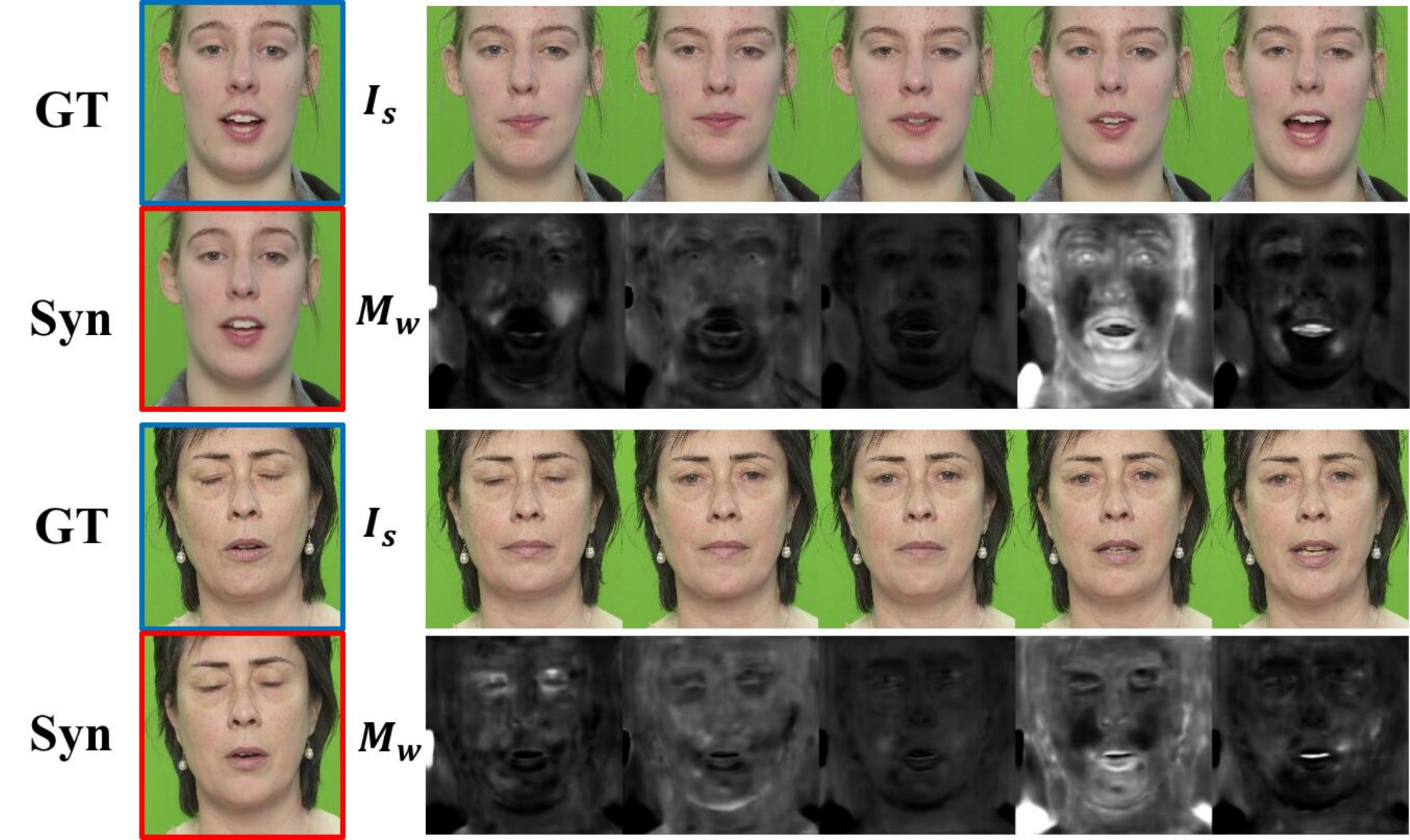}
\caption{Example of bank images selection mask $M_{w}$. Blue box: ground truth. Red box: synthesized image. Upper row right of blue box: input bank images. Lower row right of red box: Mask for bank images. Note that the tooth region in the 5th bank image was chosen, which was warped and used to generate the synthesized image. Same for the closed eye lid region in first image of the second example.}
\label{fig:mask_bank}
\end{figure}

As shown in Figure \ref{fig:mask_bank}, we also visualize the bank images selection mask $M_{w}$ to show that the network is able to synthesize facial images by using different parts from the input images, thus can generate new facial geometry. Suppose the target landmark is open mouse with closed eyes like the third row in Figure \ref{fig:mask_bank}, whereas all the input images have opened eyes except for the first one, which has closed eyes and closed mouth. In this case, the network will select the eye region from the first image and the mouth region from other matching images. This approach gives the network great flexibility, which allows it to synthesize unseen facial geometry by automatically selecting and warping the best matching regions in the input bank images.

\subsection{Failure Cases}
\begin{figure}[ht]
    \centering
    \includegraphics[width=\columnwidth]{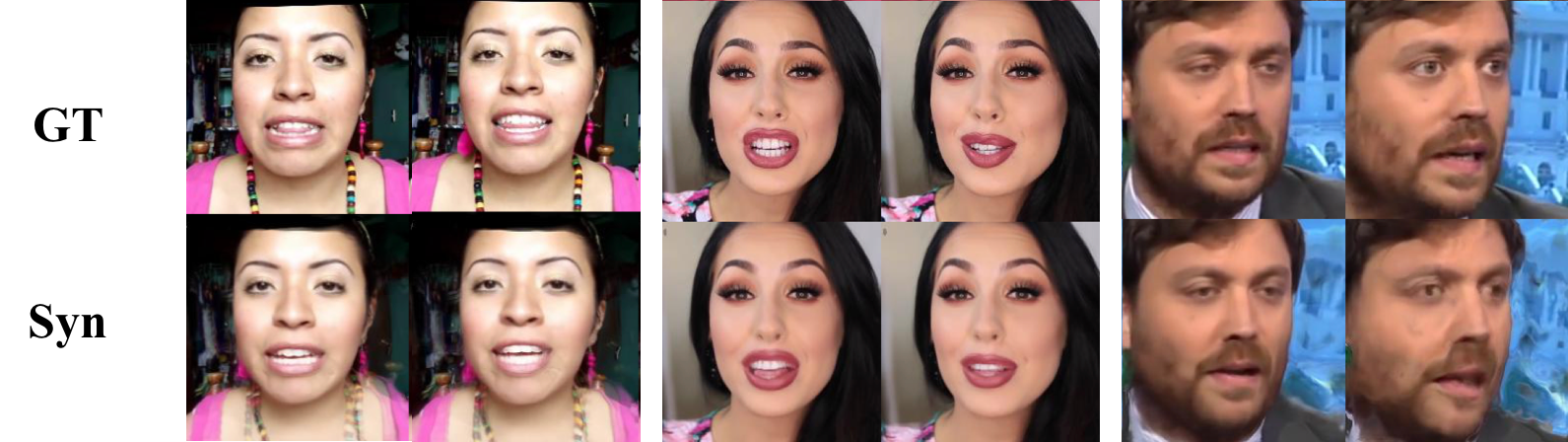}
    \caption{Failure cases: ambiguity of tooth and tongue. Left: fail to synthesize Interdental Consonant Mouth. Middle: synthesize incorrect Interdental mouth. Right: wrapped background.}
    \label{fig:fc}
\end{figure}
Due to the nature of landmark-driven methods, we found that our model fails to generate faithful mouth of Interdental Consonant in Figure \ref{fig:fc}. That's because Interdental Consonant has a quite similar mouth shape (or landmark relative positions) to other commonly-seen opening mouth. To address the issues, a multi-modal system involving audio will greatly help the system. This will be left for future work. Our model also suffers from wrapped background if the poses are too extreme.

\section{Conclusion}
In this paper, we propose a landmark driven talking facial animation synthesis system by combining appearance-based and warping-based streams. Instead of taking single source image as input, we introduce a source image bank consisting of five images with distinct landmarks. Then our warping-based fetching stream learns to select the most related facial features from the source bank to warp and merge them, and the appearance-based learning stream further compensates other unseen features from the training face feature space. The ultimately merged faces, which are highly consistent with the target faces, demonstrate more faithful synthesis performance than other baseline algorithms qualitatively and quantitatively. The advantages of our method are obvious on synthesizing tooth, eyes, or other unseen facial detailed regions. 

\section{Acknowledgments}
We would like to thank Prof. Dan Morrow and Jump ARCHES grant for their funding support during this work. In addition, we would like to thank IFP group as well as NCSA HAL cluster for providing the computational resources used in this study.

\bibliographystyle{aaai}
\bibliography{ref.bib}
\end{document}